\documentclass[letterpaper, 10 pt, conference]{ieeeconf} 

\IEEEoverridecommandlockouts 

\usepackage{cite}
\usepackage{amsmath,amssymb,amsfonts}
\usepackage{algorithmic}
\usepackage{graphicx}
\usepackage{physics}
\usepackage{caption}
\usepackage{subcaption}
\usepackage{textcomp}
\usepackage{multirow}
\usepackage{xcolor}
\usepackage{array}
\usepackage{balance}
\newcolumntype{C}[1]{>{\centering\let\newline\\\arraybackslash\hspace{0pt}}m{#1}}

\title{\LARGE \bf
NeuroSwarm: Multi-Agent Neural 3D Scene Reconstruction and Segmentation with UAV for Optimal Navigation of Quadruped Robot
}

\author{Iana Zhura, Denis Davletshin, Nipun Dhananjaya Weerakkodi Mudalige, \\ Aleksey Fedoseev, Robinroy Peter and Dzmitry Tsetserukou 
\thanks{The authors are with the Intelligent Space Robotics Laboratory, Skoltech, Bolshoy Boulevard 30, bld. 1, 121205, Moscow, Russia }
\thanks{email: \{iana.zhura, denis.davletshin, nipun.weerakkodi, aleksey.fedoseev, robinroy.peter, dzmitry.tsetserukou\}@skoltech.ru }
}

\begin{document}

\maketitle
\thispagestyle{empty}
\pagestyle{empty}

\begin{abstract}

Quadruped robots have the distinct ability to adapt their body and step height to navigate through cluttered environments. Nonetheless, for these robots to utilize their full potential in real-world scenarios, they require awareness of their environment and obstacle geometry. We propose a novel multi-agent robotic system that incorporates cutting-edge technologies. The proposed solution features a 3D neural reconstruction algorithm that enables navigation of a quadruped robot in both static and semi-static environments. The prior areas of the environment are also segmented according to the quadruped robots' abilities to pass them. Moreover, we have developed an adaptive neural field optimal motion planner (ANFOMP) that considers both collision probability and obstacle height in 2D space.

Our new navigation and mapping approach enables quadruped robots to adjust their height and behavior to navigate under arches and push through obstacles with smaller dimensions. The multi-agent mapping operation has proven to be highly accurate, with an obstacle reconstruction precision of 82\%. Moreover, the quadruped robot can navigate with 3D obstacle information and the ANFOMP system, resulting in a 33.3\% reduction in path length and a 70\% reduction in navigation time. 
The developed project with code is available on GitHub\footnote{{https://github.com/Iana-Zhura/NeuroSwarm}}.


\end{abstract}

\section{Introduction}

\begin{figure}[t]
 \centering
 \includegraphics[width=0.95\linewidth]{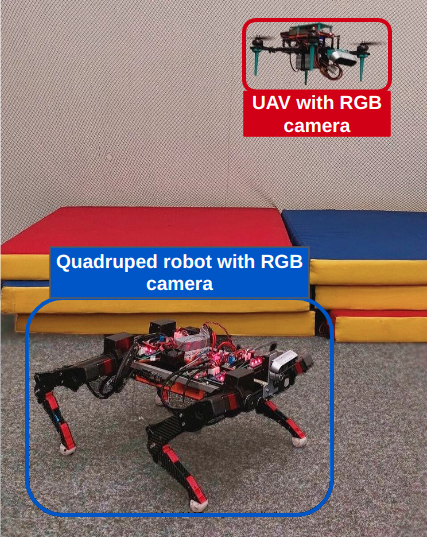}
 \caption{\centering NeuroSwarm multi-agent system.}
 \label{fig:cover_img}
 \vspace{-2em}
\end{figure}

Reflecting on the past decade, we can witness rapid progress in the field of quadruped robot locomotion and navigation \cite{market}. Quadruped robots are designed according to the biological properties of animals. With advanced control and perception systems, they can ascend stairs, jump, and even crawl; and they are expected to be able to overcome obstacles. It is anticipated that quadruped robots will be able to contribute to various missions, such as safety inspection, space exploration, and rescue operations. However, navigation in cluttered environments without being damaged or becoming trapped is quite a challenge. In fact, such missions should be undertaken without any danger of putting humans at risk.

Autonomous inspection and rescue operations remain difficult for singular robots, as their design is weighted by a high number of sensors to be able to navigate in an unstructured environment. Therefore, collaboration between several robots for navigation and localization have a high potential to improve inspection performance. UAVs have proven to be efficient assistants due to their ability to approach difficult-to-reach places \cite{Zhang}.

The mapping and segmentation of the environment is fundamental for every autonomous robot system. To that end, various sensors such as LiDAR, RGB, and depth cameras have been used. Additionally, low-cost visual methods proved to be robust and reliable for a wide scope of applications \cite{visual}, \cite{obj_detect} despite their requirement for camera pose estimation. 
Neural rendering techniques received a high surge of interest in the field of computer vision \cite{chan2022efficient}, \cite{guo2022neural}. In mapping techniques, the current focus is on neural 3D representations and rendering methods, which are implemented for the localization and navigation of autonomous robots \cite{hoeller2022neural}. However, low number of these techniques can be deployed in real-time navigation operations.

The objective of this research is to advance the field of neural 3D environment representation by considering the body and step height of a quadruped robot, developing a neural field representation of obstacles in three dimensions, and designing a multi-agent system architecture using ROS2 that integrates both a UAV and a quadruped robot (Fig. \ref{fig:cover_img}).

\section{Related Works}

\subsubsection{3D mapping techniques} 
Reconstructing the environment around a robot is a critical component of navigation and localization. The demand for accurate scene information in robotic applications has led to the emergence of various techniques. In navigation and localization, the 3D map of the environment is typically obtained using fused sensor measurements from depth sensors or LiDAR. Although these sensors are effective, they can be expensive and provide noisy data with low albedo and glossy surfaces. In contrast, image-based mapping methods are low-cost, high-resolution, and require no complex calibration. However, the current methods for 3D reconstruction utilizing monocular or multi-view techniques from RGB data can lack quality when compared to stereo methods \cite{fu2018deep, 9561077}. Moreover, scene semantic segmentation can significantly assist 3D mapping by enabling a quadruped robot to choose optimal paths through complex environments. This is particularly beneficial in cluttered terrains, such as those found on Mars, where route planning and navigation proves to be difficult \cite{Mars}.

Recent advances in Deep Learning (DL) have seen the emergence of more neural 3D reconstruction approaches, including Neural Radiance Fields (NeRFs), which replace the traditional discrete sampled geometry with a continuous volumetric function parameterized as a Multilayer Perceptron (MLP) \cite{block-nerf}. NeRFs effectively store geometry in memory as neural network weights, making them beneficial for memory-constrained systems. Nevertheless, NeRFs pre-trained model of a specific scene is not applicable for rendering other scenes in real-time, which is critical for robot operations.
In addition, Generative Adversarial Neural Networks (GAN) proved to be efficient for image enhancement in the application of visual SLAM \cite{savinykh2022darkslam}, and were also used for 3D reconstruction with additional image processing \cite{chan2022efficient}. While visual SLAM remains the most widely used technique for a 3D mapping task, it is prone to deteriorate under certain environmental conditions, as well as the presence of moving objects, which can reduce its accuracy. 

In contrast, 3D neural reconstruction based on feature extraction has the potential to provide highly precise results, as it can learn to extract general features and patterns that are applicable across different scenarios, rather than relying solely on the specific feature detection and matching techniques. This makes it a favorable alternative for applications where accuracy and robustness are the key considerations. Although an adaptive visual SLAM algorithm was prior developed \cite{orb-slam}, this approach did not allow for the incorporation of multiple sensors and synchronization, which can be essential for multi-agent systems.

\subsubsection{Quadruped robot navigation}
In the context of quadruped navigation with the collaboration of UAV, several systems were explored. Despite the fact that mapping still utilizes a 3D representation in the form of a point cloud obtained from LiDAR data, it can become sparse and fail to provide complete geometric awareness to the robot \cite{Anymal}. There are also examples of research conducted with heterogeneous multi-agent robotic systems, where UAV performs localization or navigation tasks. However, in most cases, wheeled mobile robots and UAVs are considered for collaboration  \cite{icuas1, roldan2016heterogeneous}, which limits their application. Another approach proposed a navigation of quadruped robot with human guidance through a Virtual Reality framework \cite{babataev2022hyperguider}. Although this approach has the advantage of human cognition, it does not consider the quadruped robot's capability to navigate through narrow passageways, such as arches and windows. Additionally, in areas with limited visibility, it heavily depends on human intervention for guidance.

In this paper, we present a novel multi-agent robotic system NeuroSwarm in which a mini-UAV assists the quadruped robot HyperDog to collect RGB data of the environment for neural 3D scene mapping and prior areas segmentation. The obtained 3D map is utilized by the adaptive neural field optimal motion planner for quadruped robot. This unique system highlights the benefits of combining state-of-the-art techniques in multi-agent robotics \cite{de2022marsupial, adamkiewicz2022vision, mildenhall2021nerf}.

 \section{Overview of NeuroSwarm System}

\begin{figure}[h!]
 \centering
 \includegraphics[width=0.95\linewidth]{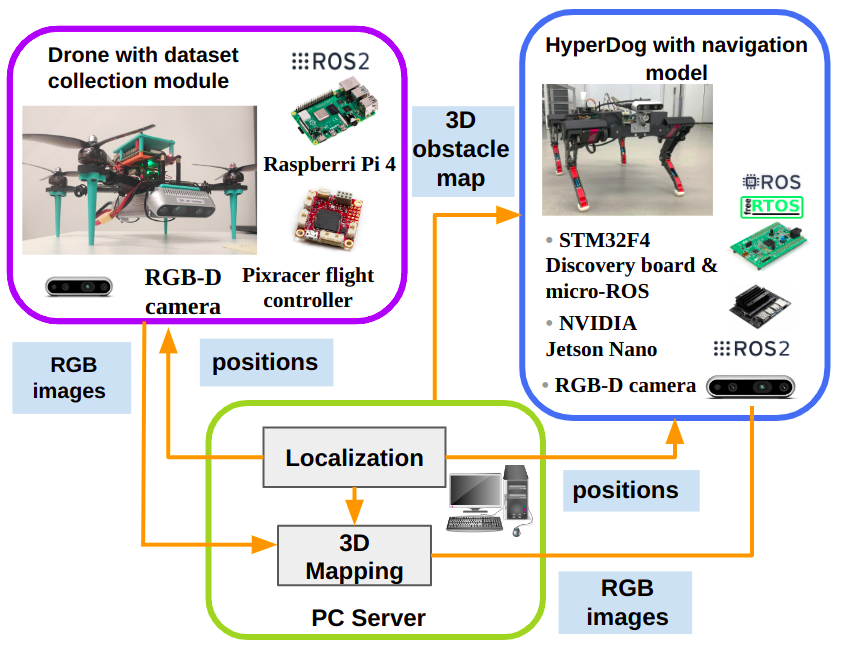}
 \caption{\centering System overview of NeuroSwarm hardware.}
 \label{fig:hardware}
\end{figure}

The hardware system consists of the quadruped robot HyperDog \cite{Nipun}, an UAV, and a personal computer (PC) with Intel Core i7 processor and Ubuntu OS. The PC server has two modules: localization and 3D mapping. HyperDog is equipped with an onboard computer NVIDIA Jetson Nano, a microcontroller STM32F4, an inertial measurement unit (IMU), and a battery pack. The drone utilizes a Raspberry Pi 4 (4GB RAM), and Pixracer R14 as a flight controller. The drone is equipped with 14.8V and 2000mAh power supply, and provides 15 minutes of flight. Both HyperDog and the UAV have Intel RealSense D435i cameras (Fig. \ref{fig:hardware}) used as an RGB-D image source excluding depth data.

\begin{figure}[h!]
 \centering
 \includegraphics[width=0.93\linewidth]{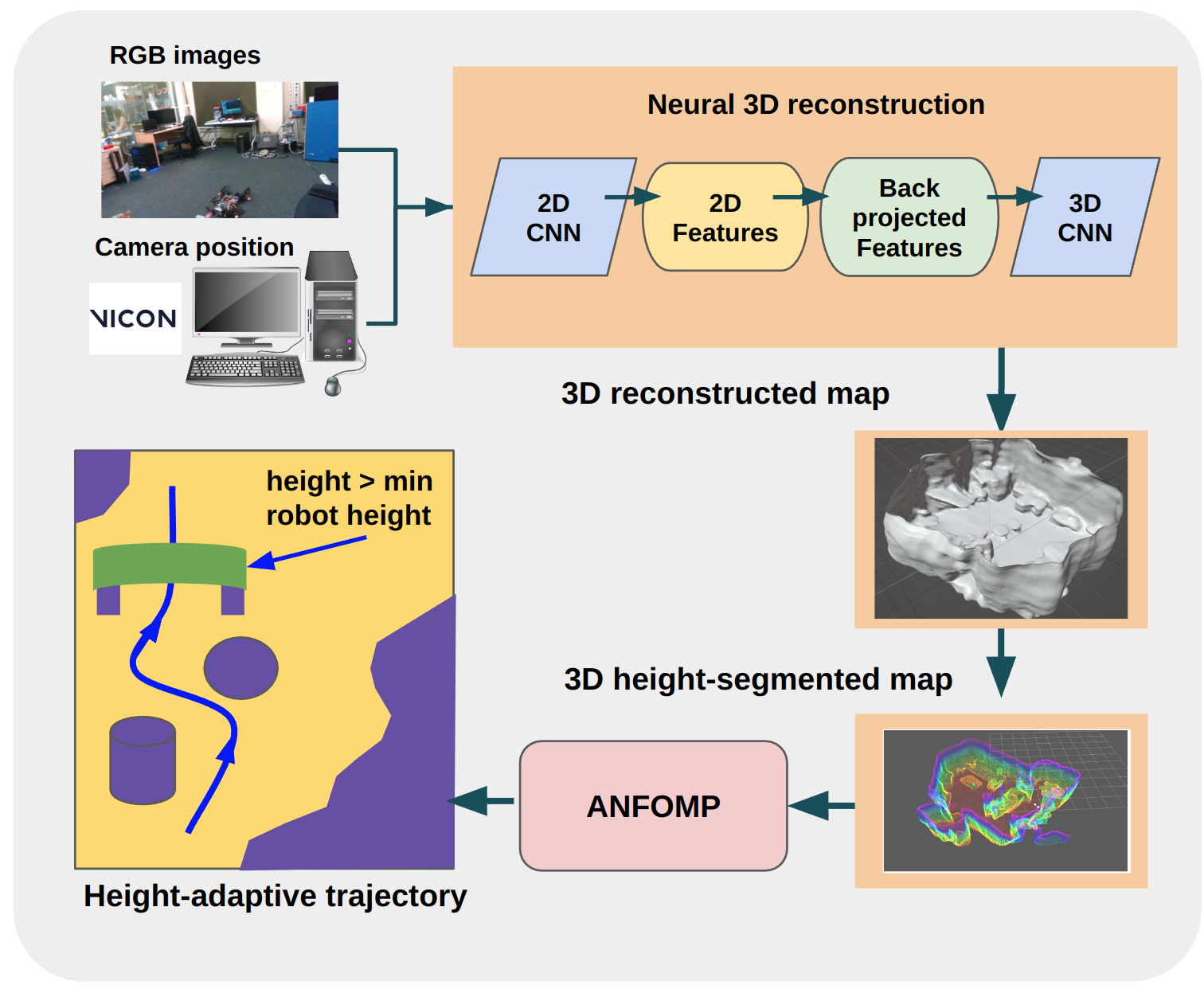}
 \caption{System architecture of NeuroSwarm software. Image frames and camera positions are stored with the ROS2 framework. The 3D map is acquired using the neural 3D reconstruction technique, which is subsequently processed to generate a comprehensive 3D height-segmented map.}
 \label{fig:software}
\end{figure}

There are three main modules in the software architecture of the multi-agent system, which include neural 3D reconstruction, 3D height-segmented obstacle map extraction and adaptive neural field optimal motion planner (ANFOMP).
The software architecture is presented in Fig. \ref{fig:software}.

The 3D neural reconstruction model processes each camera frame to generate a 3D map, which is then used to create a 3D height-segmented obstacle map. This helps to identify the passable objects for the quadruped robot in the environment, allowing it to navigate safely and accurately.

\subsection{3D Reconstruction}
The proposed 3D reconstruction algorithm utilizes ATLAS \cite{murez2020atlas} which integrates 2D Convolutional Neural Networks (CNNs) for feature extraction. These extracted features are then back projected into voxel volumes by factoring in camera positions and intrinsic camera parameters. The voxel volumes are further refined using 3D CNNs, and the environment is represented as a 3D mesh with predicted segmentation. This baseline algorithm has been enhanced with a 3D map to extract a new 3D height-based segmented obstacle map based on the quadruped robot height, enabling better understanding of the terrain and enhancing navigation.

\begin{figure}[h!]
 \centering
 \includegraphics[width=0.88\linewidth]{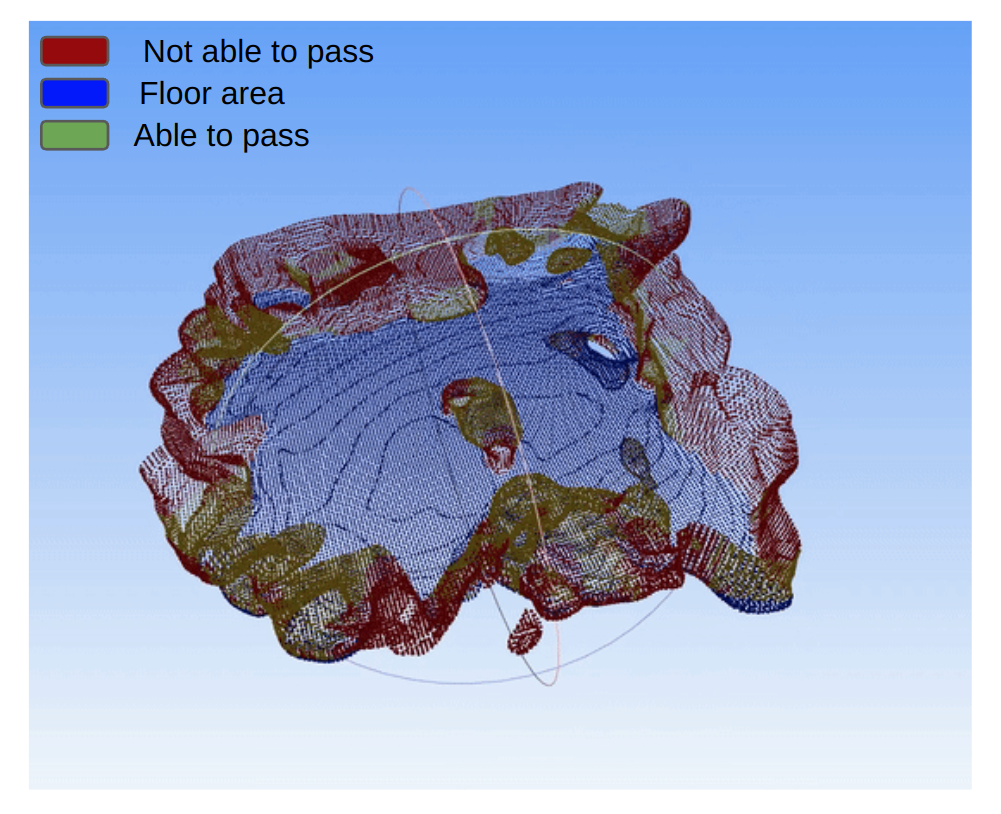}
 \caption{The 3D obstacle map segmented based on height. The green voxels represent passable areas with a height less than the maximum robot height.}
 \label{fig:3D_obs_map}
 \vspace{-1em}
\end{figure}

To generate a 3D height-segmented obstacle map, it is necessary to determine the floor plane height and the distance of obstacles from it, at each vertex. This data is then compared against the parameters of the quadruped robot, such as its maximum step height, maximum and minimum heights. As shown in Fig. \ref{fig:3D_obs_map}, the 3D height-segmented obstacle map obtained from a dataset collected collaboratively reveals that objects can be distinguished and identified separately from the floor area. 

\subsection{Adaptive Neural Field Optimal Motion Planner}
To validate the navigation ability of the quadruped robot in the environment utilizing the proposed mapping method, neural field optimal motion planner (NFOMP) was selected as the baseline algorithm \cite{Kurenkov}.

In comparison with classical sampling-based path planning algorithms, such as RRT, RRT\({^*}\) \cite{RRT} which have been widely applied for the quadruped robot navigation, NFOMP allows for achieving a smooth and optimal path in a reasonable amount of time. NFOMP represents one of the recently introduced path planning algorithms, aimed at addressing the issues of jagged and shortcut trajectory, as well as U-shaped obstacles. This algorithm employs online neural field training during trajectory optimization to predict the probability of a collision at any given position.

The NFOMP framework utilizes a collision model, where NeRFs are employed to efficiently store the geometry of obstacles in a 2D space. However, NFOMP has currently only been employed to carry out 2D trajectory planning in static environments for differential drive robots.

In this research, we have advanced the capabilities of the NFOMP for navigating in a 3D environment with variable height. We have extended the collision model to include the third dimension of obstacles and integrated NFOMP with 3D height-segmented obstacle mapping and trajectory generation only in regions where HyperDog can move at maximum or minimum height. 

\section{Heterogeneous Mapping Pipeline}

\begin{figure}[h!]
 \centering
 \includegraphics[width= 0.95\linewidth]{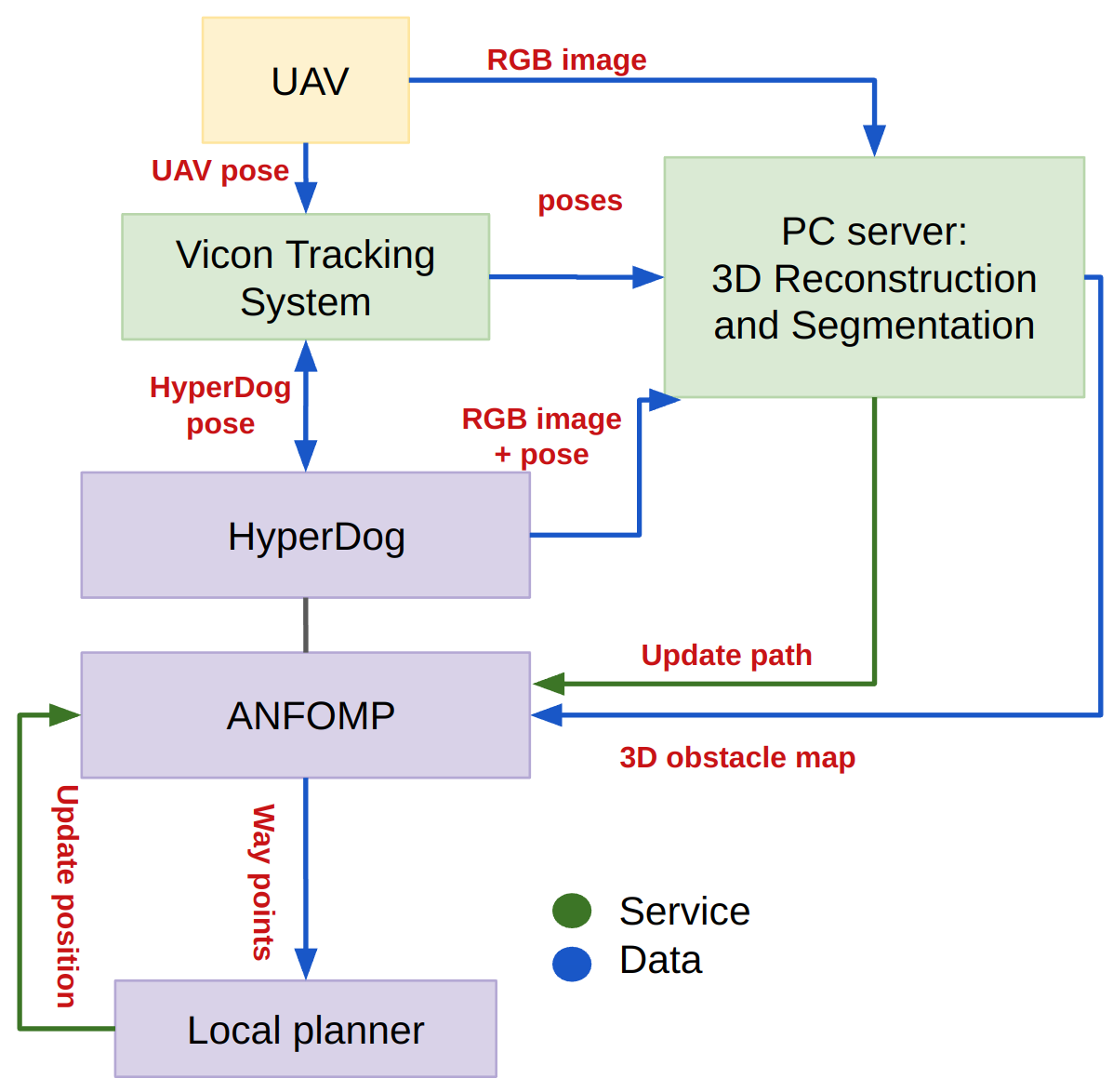}
 \caption{Pipeline of multi-agent cooperative navigation.}
 \label{fig:multiagent}
 \vspace{-1em}
\end{figure}

To generate the map through a collaboration of two robots, images with their respective positions must be captured and processed. The 3D map can be updated, when it is needed for motion planner. The multi-agent operation pipeline is demonstrated in Fig. \ref{fig:multiagent}. 

The ROS2 framework transfers images and positions to the PC server, where the dataset from the UAV and HyperDog is processed using a neural 3D reconstruction network and 3D height-segmented obstacle map algorithm. The ANFOMP algorithm of HyperDog uses the 3D obstacle map to generate an optimal trajectory while accounting for obstacle height. The map can be updated incrementally, allowing for close to real-time recalculation and path optimization for the quadruped robot.

\section{Experiments}
We conducted the experiments in a designated room measuring 5x5 meters, equipped with the Vicon V5 Mocap System that offers sub-millimeter precision in localization. To construct a map, we guided the robots along predetermined trajectories that ensured complete coverage of the environment for reconstruction.

\subsection{Dataset Collection}
To achieve 3D scene reconstruction, we collected a collaborative dataset utilizing Intel RealSense D435i cameras installed on both the UAV and HyperDog. The positions of both robots were determined by the Vicon Tracker System. We devised a ROS2 framework for a multi-agent system to facilitate the exchange of information between the robots, enabling the transmission of image captures and position data every second. The dataset consists of 250 photos of the environment, which were collected in two distinct sets. The first set includes both the UAV and HyperDog, while the second set solely focuses on the quadruped robot.

\subsection{First Experiment: Map Evaluation}

In the first experiment, our objective was to evaluate the quality of the map in two scenarios: when images were collected solely by the quadruped robot and when images were simultaneously collected by the UAV and HyperDog. The algorithm implemented in the ROS2 framework efficiently processed the dataset within a time frame of 90 seconds. During this period, the UAV and the quadruped robot successfully captured 250 images from various perspectives, while HyperDog alone, limited by the height of the quadruped robot, collected 80 images from the specific viewpoints.

\begin{figure}[h!]
 \centering
 \includegraphics[width=\linewidth]{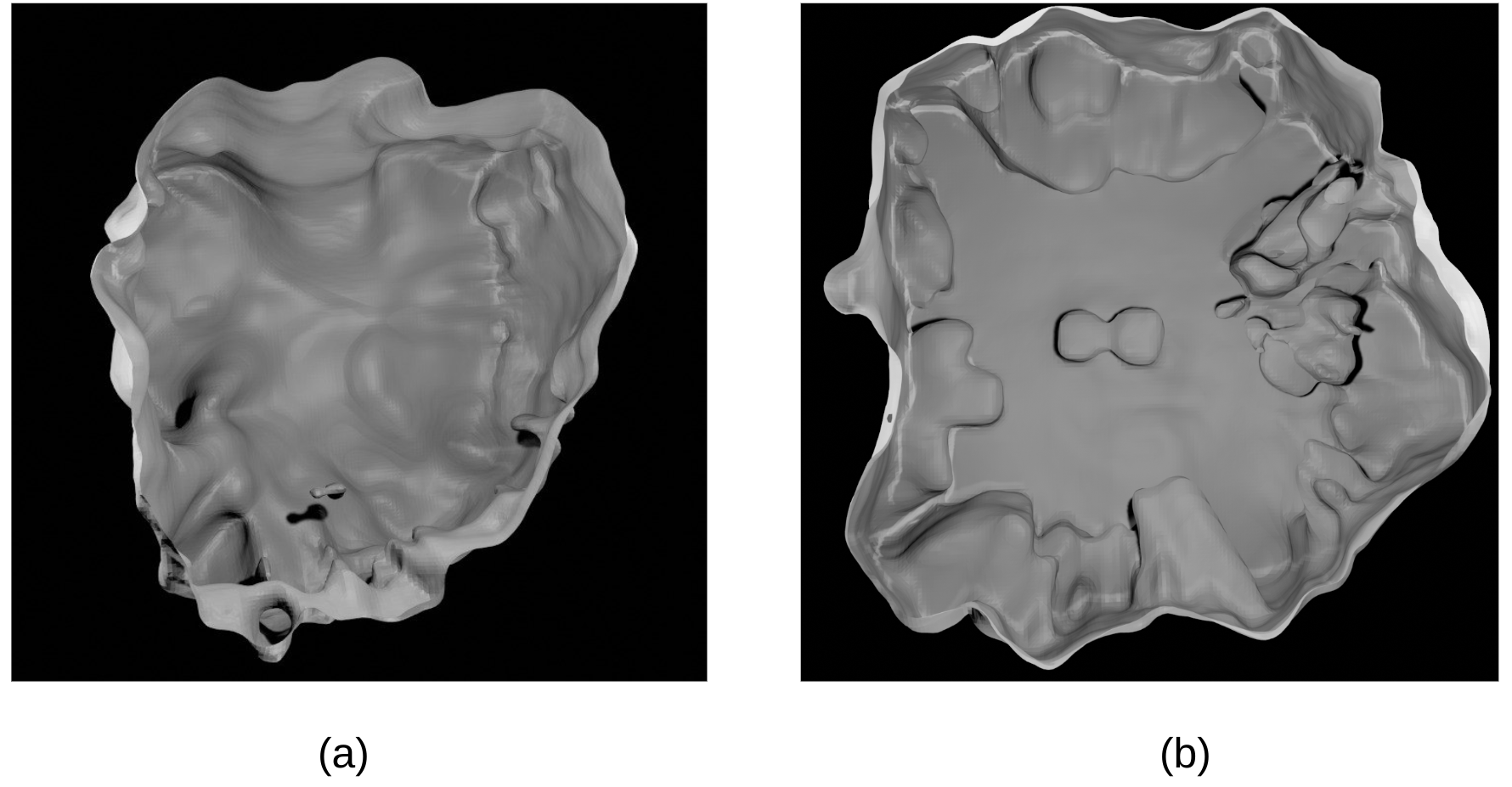}
 
 \caption{Top view of obtained map. (a) From HyperDog. (b) From both the UAV and HyperDog.}
 \label{fig:experiment_1}
\end{figure}

Fig. \ref{fig:experiment_1}(a) illustrates that the map reconstructed from the HyperDog dataset lacked the distinctive features present in the experimental room. This poses a challenge for quadruped robot navigating in cluttered environments, as it is crucial to differentiate objects from the floor plane to adjust the robot's height accordingly. In contrast, the map derived from the collaborative dataset encompassed a variety of objects, including those with passages (Fig. \ref{fig:experiment_1}(b)).

It is evident that objects can be easily distinguished from the floor area and recognized as separate entities. To further enhance the accuracy of our mapping, we integrated VICON indoor localization system for odometry.

\begin{figure}[h!]
 \centering
 \includegraphics[width=0.9\linewidth]{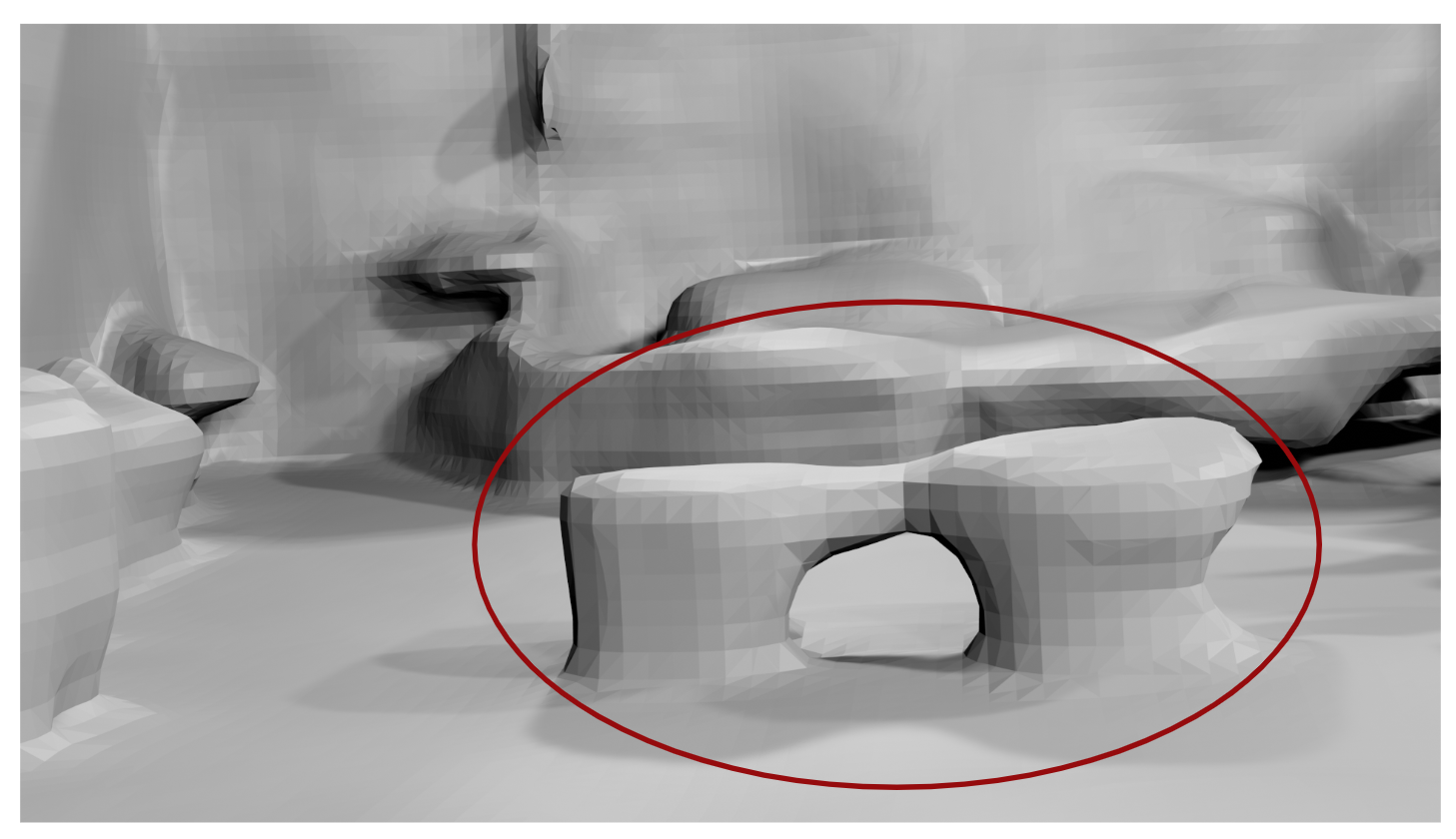}
 \caption{Frontal view of collaborative map: The passages through and under the objects have been reflected in the map.}
 \label{fig:exp_1_arch}
\end{figure}

In order to evaluate the quality of the map, we calculated geometry metrics for passable objects commonly encountered by quadruped robots, such as tables, arches, and chairs (see Fig. \ref{fig:exp_1_arch}). Precision was assessed using the Intersection over Union (IoU) metric, which compares the bounding boxes of the ground truth with those of the 3D reconstructed model. By applying the IoU metric and the volumes of the bounding boxes, we were able to construct a confusion matrix that compared the predicted and actual dimensions. This allowed us to estimate metrics such as Recall, Precision, and F-score. The ground truth point cloud data was collected through the use of LiDAR.

\begin{figure}[h!]
 \centering
 \includegraphics[width=0.8\linewidth]{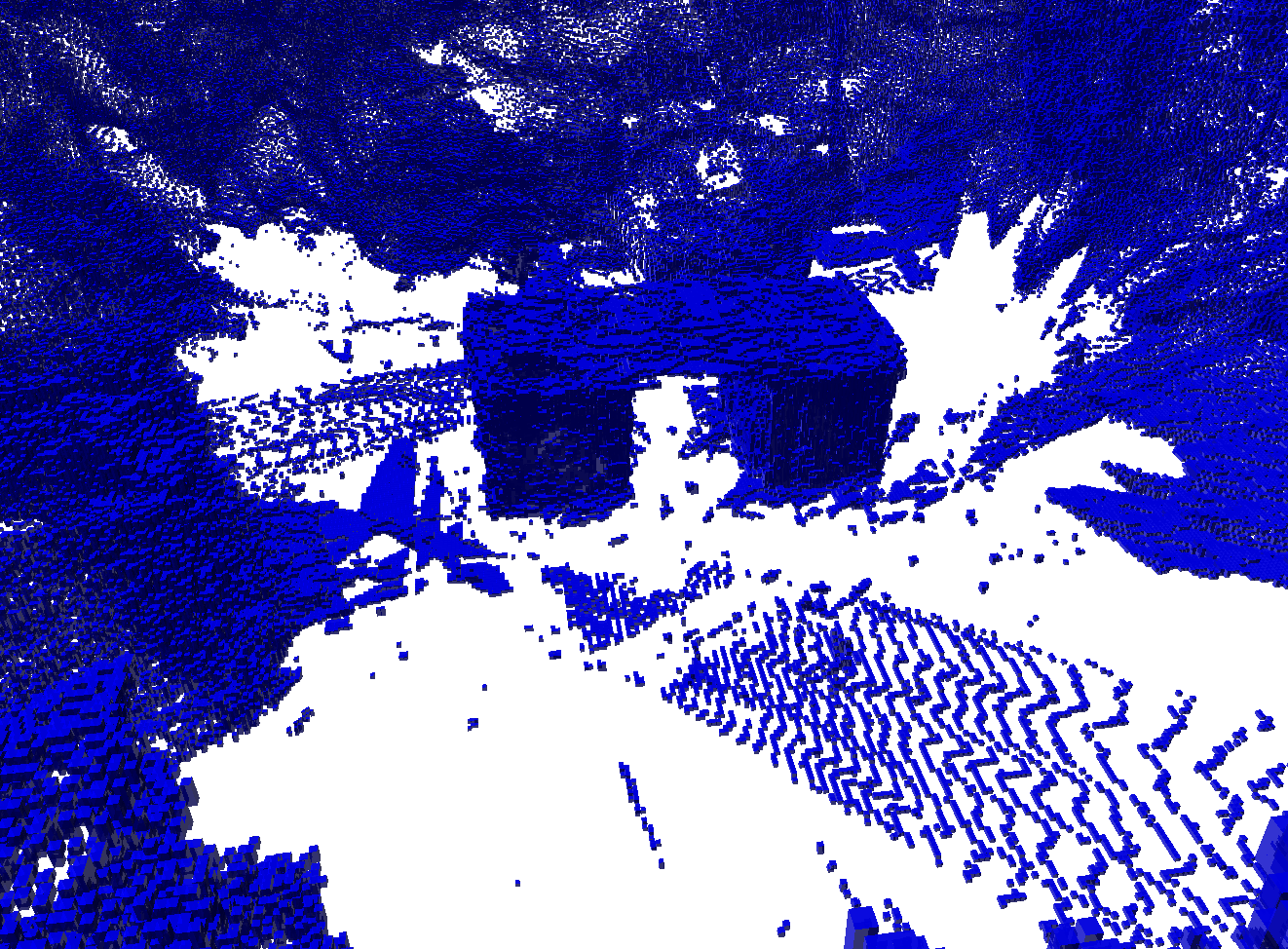}
 \caption{Map obtained via OctoMap algorithm, which employs the RealSense D435i RGB-D in ROS2.}
 \label{fig:exp_1_octomap}
 \vspace{-1em}
\end{figure}

To compare our proposed system with existing alternatives, we utilized the OctoMap algorithm, which employed the RealSense D435i RGB-D in ROS2, specifically making use of the RealSense2-camera and OctoMap-server2 packages (Fig. \ref{fig:exp_1_octomap}). 

To assess the shape of the reconstructed passable object, we calculated the root mean squared error (RMSE) of the signed distance function (SDF) for each point of the reconstructed object in relation to the ground truth point cloud.

Table \ref{table_metrics} demonstrates that our 3D reconstruction mapping approach, when compared to a classical depth mapping technique, allowed us to generate a more detailed and accurate model of the environment. 

\begin{table}[!h]
\caption{Geometry Metrics of Passable Objects and Mapping Time.}
\label{table_metrics}
\begin{center}
\begin{tabular}{|C{3cm}|C{2cm}|C{2cm}|}
\hline
\multirow{2}{*}{Metrics} & \multicolumn{2}{C{4cm}|}{Method}\\
\cline{2-3}
& Atlas & OctoMap\\
\hline
Precision & \textbf{0.82} & 0.68 \\
\hline
Recall & \textbf{0.98} & 0.91 \\
\hline
F-score & \textbf{0.86} & 0.7 \\
\hline
Time, sec &  \textbf{30} & 240\\
\hline
RMSE SDF, m & \textbf{0.002}& 0.007\\
\hline
IoU, \%  & \textbf{83}& 75\\
\hline
\end{tabular}
\end{center}
\end{table}
\vspace{-1em}

With a precision of 82\%, this approach proved to be suitable for our research purposes. The OctoMap algorithm takes approximately 240 seconds to generate a comprehensive 3D map. In contrast, a height-segmented obstacle map can be derived from a mesh in less than 2 seconds. Additionally, it is important to highlight that the OctoMap algorithm does not offer segmentation for objects that are passable.



\subsection{Second Experiment: Motion Planner Validation}
\begin{figure}[h!]
 \centering
 \begin{subfigure}[b]{0.478\linewidth}
 \centering
 \includegraphics[width=\textwidth]{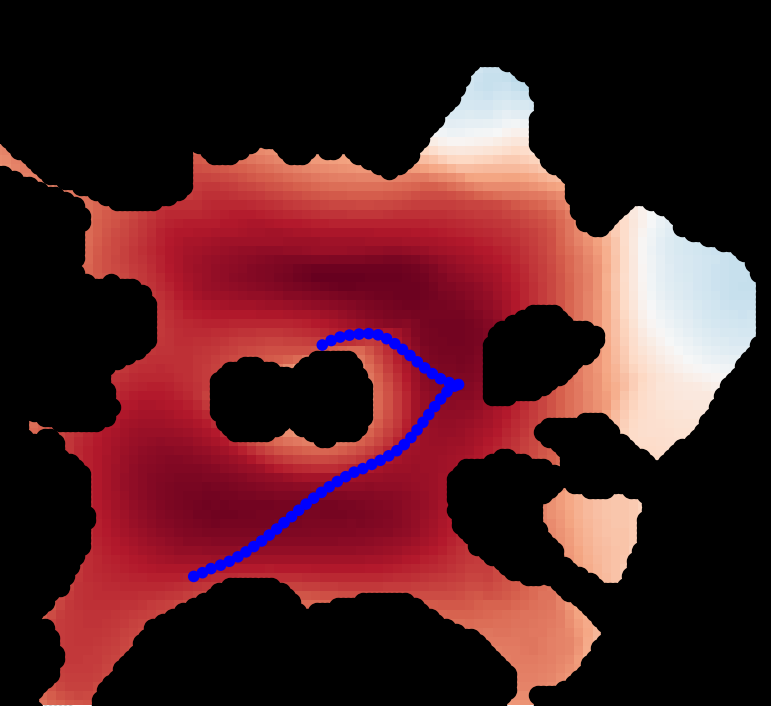}
 \caption{}
 \label{fig:2D_traj}
 \end{subfigure}
 \begin{subfigure}[b]{0.49\linewidth}
 \centering
 \includegraphics[width=\textwidth]{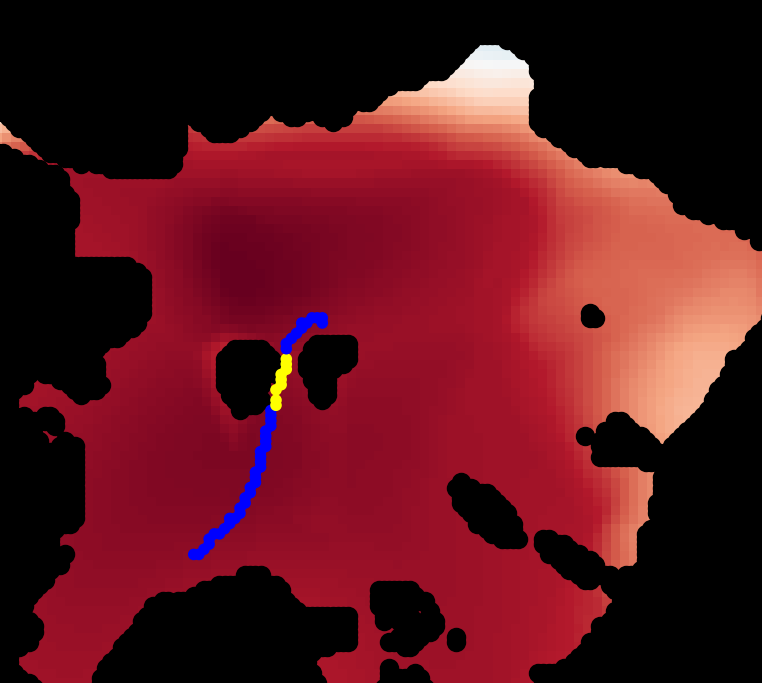}
 \caption{}
 \label{fig:3D_traj}
 \end{subfigure}
 \caption{(a) The trajectory generated using 2D obstacle information and NFOMP. (b) The trajectory obtained via 3D height-segmented obstacle map and ANFOMP: The trajectory part, where robot must change height, is marked with yellow color, blue trajectory indicates path with maximum height. }
 \label{fig: experiment_2}
\end{figure}

The second experiment aimed to generate two trajectories for the quadruped robot using two types of obstacle maps. In the first case, only 2D obstacle information was considered for trajectory generation, which involved the application of the 2D NFOMP proposed by Kurenkov et al. \cite{Kurenkov}. The trajectory in this case was designed to bypass the frame around the obstacles, which proved to be time-consuming (Fig. \ref{fig: experiment_2}(a)).

In Fig. \ref{fig: experiment_2}, the red fields on the map indicate a low collision probability, while the blue fields represent a high collision probability. The areas where the robot cannot traverse due to its physical limitations are marked in black.
 
In the second case of the experiment, a trajectory was generated by incorporating both 2D and 3D obstacle information. This trajectory was designed to navigate through the frame based on the 3D dimensions of the obstacles, with ANFOMP receiving information about the necessary amount of ducking for the quadruped robot to pass through. Fig. \ref{fig: experiment_2}(b) illustrates that the planned trajectory passes under the arch, with yellow points indicating the specific areas where the robot must adjust its height to the minimum required for traversal. Each point along the generated trajectory includes height information for the quadruped robot. This approach represents an improvement over the first case, which solely considers 2D obstacle information \cite{Kurenkov}.

Table \ref{table_nav} shows the results of trajectory length estimations for 2D trajectory generation and for the new height-adaptive algorithm. 
 \begin{table}[h!]
\caption{Comparison of 2D and HAT Trajectories.}
\label{table_nav}
\begin{center}
\begin{tabular}{|c|c|c|c|}
\hline
Metrics & Max curv. & Path Length, m & Time, sec \\
\hline
2D trajectory & 16 & 5.2 & 200 \\
\hline
HAT trajectory & \textbf{5.4} & \textbf{3.6} & \textbf{60}\\
\hline
\end{tabular}
\end{center}
\end{table}
The height-adaptive technique (HAT) proved to be more effective than 2D path planning for quadruped robot navigation, resulting in reduced time to reach the target without necessitating drastic turns. However, this improvement can only be achieved with the assistance of the UAV, which can swiftly collect and map the environment with high precision in 95 seconds, compared to the 8 minutes required by the quadruped robot alone. Furthermore, the first experiment demonstrated that the quadruped robot, operating solely on the distinguishable objects, was unable to generate a map within a timeframe comparable to that of a multi-agent system. Taking into account the quadruped robot's locomotion speed of 0.2 m/s towards the target, the entire operation lasted 120 seconds. The deliberately reduced speed allowed the quadruped robot to capture images during its journey.

\section{Conclusions and Future Work}

Our project proposes a novel multi-agent robotic system, NeuroSwarm, that enables quadruped robots to navigate in cluttered environments using the UAV for mapping and wide-area observation. We have developed navigation algorithms with 3D reconstruction and adaptive motion planner. The multi-agent mapping operation achieved an obstacle reconstruction precision of 82\%, providing accurate 3D obstacle information for the quadruped robot. The adaptive neural field optimal motion planner (ANFOMP) took into account both collision probability and obstacle height in 2D space, resulting in a 33.3\% reduction in path length and a 70\% reduction in navigation time compared to traditional sampling-based algorithms.
By incorporating the proposed multi-agent system, quadruped robots can effectively maneuver through complex terrains, overcoming obstacles and adapting their behavior to the environment.

In the future we are planning to further improve the developed system. The experiment will be conducted, where the 3D map is updated and a height-adaptive trajectory is continuously re-planned in near real-time. Furthermore, the computational requirements are such that the 3D mapping algorithm can be transferred from a PC server to onboard computers on the robot. Moreover, the multi-agent system architecture has been developed to accommodate the contribution of multiple UAVs in environment mapping.









\addtolength{\textheight}{-20cm}
\balance

\bibliographystyle{IEEEtran}
\bibliography{References.bib}

\end{document}